\documentclass[11pt]{article}
\usepackage[final]{rbs}

\usepackage{amsmath,amsfonts,bm}
\usepackage{xspace}
\usepackage{xcolor}

\def\eqref#1{equation~\ref{#1}}
\def\1{\bm{1}}

\DeclareMathAlphabet{\mathsfit}{\encodingdefault}{\sfdefault}{m}{sl}
\SetMathAlphabet{\mathsfit}{bold}{\encodingdefault}{\sfdefault}{bx}{n}

\usepackage{hyperref}
\usepackage{url}
\usepackage{graphicx}
\usepackage{float}
\usepackage{booktabs}
\usepackage{multirow}
\usepackage{subcaption}
\usepackage{wrapfig}
\usepackage{xcolor}
\usepackage{xspace}
\usepackage{amsmath}
\usepackage{enumitem}
\usepackage{booktabs}
\usepackage{multirow}
\usepackage{array}

\newcommand{\modelname}{\text{GenSight}\xspace}

\title{Seeing and Imagining as One: Unified Video Generation and 4D Perception for Embodied World Modeling}

\author{
\hypersetup{hidelinks}
\bfseries Haitang Feng$^{1,2,*}$, \href{https://zhuomanliu.tech/}{Zhuoman Liu}$^{1,3,*}$, \href{https://scholar.google.com/citations?user=fDUwF94AAAAJ&hl=zh-CN}{Zhennan Wang}$^{1,\dagger}$,\\ \href{https://chantsss.github.io/}{Sheng Wang}$^{1,\dagger}$, Juelin Li$^{1}$, Chengzhi Zhao$^{1}$, Hexiang Wei$^{1}$, \href{https://linsats.github.io/}{Lin Shao}$^{1,4,\ddagger}$\\[0.35em] \normalfont\small $^{1}$\href{https://en.roboscience.co/}{RoboScience} \quad $^{2}$Nanjing University\\[0.15em]
$^{3}$The Hong Kong Polytechnic University \quad $^{4}$National University of Singapore}

\NewCommandCopy\rbslogooriginal\rbslogo
\renewcommand{\rbslogo}[1][black]{\begingroup\hypersetup{hidelinks}\href{https://en.roboscience.co/}{\rbslogooriginal[#1]}\endgroup}

\begin{document}

\maketitle
\begingroup
\makeatletter
\renewcommand{\@makefntext}[1]{\noindent#1}
\makeatother
\renewcommand{\thefootnote}{}
\renewcommand{\theHfootnote}{rbscredits}
\begin{NoHyper}
\footnotetext{\textsuperscript{*} Equal contribution.\quad \textsuperscript{\ensuremath{\dagger}} Project lead.\quad \textsuperscript{\ensuremath{\ddagger}} Corresponding author.}
\end{NoHyper}
\endgroup

\begin{abstract}
Embodied world models should not only imagine plausible futures, but also perceive the geometric, semantic, and motion structure of the evolving world.
We present \modelname{}, a unified embodied world model that brings future video generation and dense perception into a single pretrained video diffusion transformer.
\modelname{} uses a unified spatiotemporal representation for video, depth, embodiment segmentation, and first-frame-relative 2D flow.
Built on a shared DiT backbone, \modelname{} uses lightweight task modulation to predict all modalities within the same architecture.
All tasks are jointly optimized within a unified velocity prediction space.
At inference, \modelname{} supports two complementary modes: multi-step future video generation with aligned dense predictions and single-step dense perception from an observed video. 
By unifying modality representation, model architecture, and prediction space, GenSight avoids separate modality-specific pipelines and couples appearance, geometry, semantics, and motion in a shared representation. 
It promotes cross-modal consistency and allows dense perception to benefit from the spatiotemporal priors learned through video generation. 
Trained on embodied manipulation data spanning multiple robot embodiments and human-hand interactions, \modelname{} demonstrates strong performance in both video generation and dense 4D perception, highlighting the effectiveness of unifying seeing and imagining for embodied world modeling.
{\sffamily\hypersetup{hidelinks}
\href{https://gensight-rbs.github.io/}{\textcolor{rbsGray}{\raisebox{-1pt}{\setlength{\unitlength}{0.7pt}\begin{picture}(10,10)\linethickness{0.35pt}\put(5,5){\circle{10}}\put(5,5){\oval(4,10)}\put(0,5){\line(1,0){10}}\end{picture}}}\hspace{0.15em}\textbf{Project page:}\enspace\textcolor{rbsGray}{\textsf{gensight-rbs.github.io}}}}
\end{abstract}

\begin{figure}[H]
\centering
% \vspace{-0.25cm}
\includegraphics[width=1.0\linewidth,trim=20pt 0pt 0pt 0pt,clip]{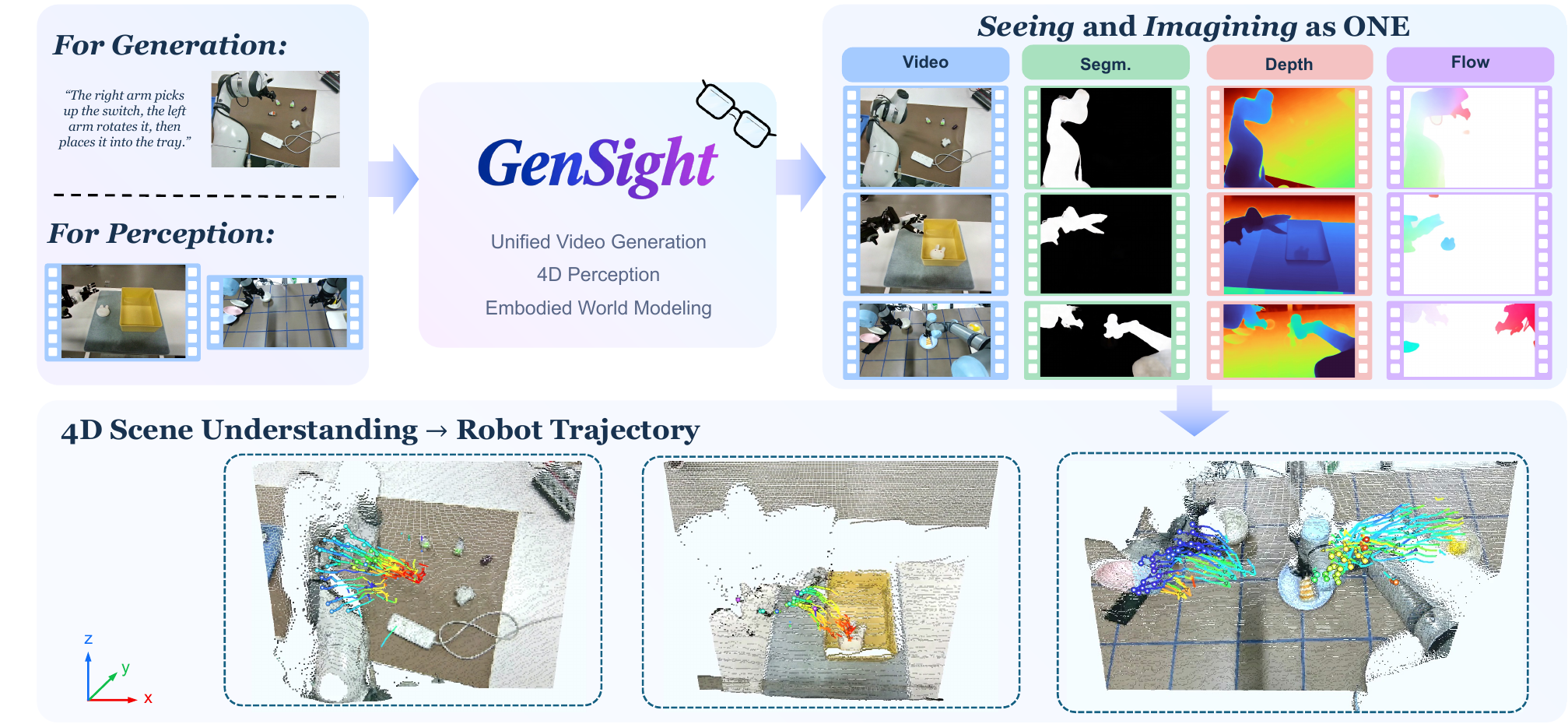}
% \vspace{0.05cm}
\caption{\textbf{\modelname{} unifies seeing and imagining for embodied world modeling.}
Given an initial image and a text prompt, \modelname{} generates future video together with aligned embodiment segmentation, depth, and first-frame-relative 2D flow.
Given an observed video, the same model predicts each dense modality in a single step.
Together, these outputs characterize appearance, semantics, geometry, and motion.
The bottom panel illustrates 4D scene understanding through scene geometry and robot motion trajectories over time.}
\label{fig:teaser}
\end{figure}

\section{Introduction}
\label{sec:intro}

Large-scale video generators have learned spatiotemporal priors~\citep{king2026gen4u,rw_wang2026genception} and can synthesize realistic future video from images and language~\citep{wan2025wan22,nvidia2026cosmos3}. 
However, for embodied interaction, RGB video is only one view of the predicted world and does not directly provide structured state estimates needed to understand the environment, analyze agent-environment dynamics, or support reasoning~\citep{guo2026unified4dworldaction,zhao2026rynnworld4d4dembodiedworld}. 
Combining these capabilities allows an embodied world model to \emph{see} the structured state and \emph{imagine} how it may evolve, providing a unified basis for understanding embodied dynamics.

Despite recent efforts to connect generation and perception, several limitations remain in embodied world modeling. 
First, future generation and dense perception are commonly developed as separate systems: world models generate RGB futures, while independent estimators recover their physical and semantic structure~\citep{chen2025large,yang2024depthanythingv2,ravi2025sam2,teed2020raft}. 
This generate-then-estimate pipeline can propagate visual artifacts into perceptual outputs and offers no mechanism for cross-modal constraints during training. 
Second, attempts to incorporate structured signals commonly rely on modality-specific branches or heads~\citep{guo2026unified4dworldaction,yuan2026dreamwam}, or parallel-stream architectures~\citep{zhao2026rynnworld4d4dembodiedworld,chen2026flowwamopticalflowunified}, increasing architectural fragmentation. 
Moreover, these cues are used mainly as auxiliary supervision, rather than enabling the same model to densely perceive observed interactions. 
Third, although generative models have been adapted for dense perception~\citep{rw_wang2026genception,rw_guo2026unigp,zhang2026dvd}, jointly supporting future video generation and dense perception through a fully shared model with a common prediction formulation remains underexplored. 
This raises a central question: can a single embodied world model both imagine futures and perceive interactions? 

To this end, we introduce \modelname{}, a unified embodied world model built on a pretrained Video DiT. 
\modelname{} unifies video, depth, embodiment segmentation, and first-frame-relative 2D flow within a common spatiotemporal latent representation, all processed by the same frozen video autoencoder. 
A shared backbone predicts each modality through lightweight task modulation, while the Transformer blocks and decoder remain fully shared, without modality-specific output heads. 
Beyond unifying representations and architecture, all tasks are jointly learned in a unified velocity prediction space, aligning dense perception with the video generator's native prediction formulation. 
The model supports two complementary inference regimes: multi-step future video generation with aligned dense readouts, and single-step dense perception from an observed video. 

This three-level unification yields three important advantages. 
First, modeling all modalities within a shared representation promotes temporal and cross-modal consistency, reducing the mismatch inherent in post-hoc perception pipelines. 
Second, joint learning enables reciprocal transfer across tasks: the rich spatiotemporal priors of video generation support dense perception, while supervision from geometry, semantics, and motion encourages the shared representation to capture physically meaningful structures. 
Third, unifying the model and prediction space avoids the need for modality-specific pipelines while preserving operational flexibility: the same model can imagine future appearances or perceive an observed interaction under different inference regimes. 
Together, these properties transform a pretrained video generator into a unified embodied world model that jointly models multiple facets of embodied dynamics. 

Our contributions are summarized as follows:
 \vspace{-2mm}
\begin{itemize}[leftmargin=*, labelindent=0pt]
    \item We study the task-level unification of future video generation and dense perception for embodied world modeling, enabling a single model to both imagine future observations and infer structured states from observed videos. 
    \vspace{-0.5mm}
    \item We introduce \modelname{}, which achieves three-level unification across heterogeneous modalities through a common spatiotemporal latent representation, a fully shared pretrained Video DiT without modality-specific output heads, and a unified velocity prediction space. 
    \vspace{-0.5mm}
    \item We validate \modelname{} across diverse robot embodiments, human-hand interactions, and unseen tasks, demonstrating strong performance in both generation and perception and providing a systematic empirical analysis of our design choices.
\end{itemize}

\section{Related Work}
\label{sec:related}

\textbf{Video generation and embodied world models.}
Video diffusion models, including Stable Video Diffusion~\citep{blattmann2023svd} and Wan 2.2~\citep{wan2025wan22}, provide spatiotemporal priors for conditional generation. 
Cosmos3-Nano extends video generation within an omnimodal world model~\citep{nvidia2026cosmos3}.
Recent work adapts video generation to embodied world modeling by predicting future observations conditioned on embodied interactions, thereby providing a generative model of environment dynamics for robot planning and control~\citep{du2023unipi,yang2023unisim}. 
RynnWorld-4D~\citep{zhao2026rynnworld4d4dembodiedworld} jointly generates RGB, depth, and optical flow using modality-specific Transformer branches.
In contrast, \modelname{} shares the Transformer backbone across modalities and supports dense perception of observed videos.

\textbf{World action models.}
World action models couple future prediction with robot control. 
X-WAM and FlowWAM incorporate depth and optical flow, respectively~\citep{guo2026unified4dworldaction,chen2026flowwamopticalflowunified}, while Flex-$\pi$ jointly models RGB, geometry, semantic features, and actions~\citep{yan2026flexpi}.
EgoWAM studies alternative visual prediction targets for human-to-robot transfer~\citep{li2026egowamworldactionmodels}, and WAM4D and DreamWAM use auxiliary visual supervision with the corresponding branches removed at deployment~\citep{li2026wam4dfast4dworld,yuan2026dreamwam}. 
Rather than directly predicting robot actions, \modelname{} unifies future video generation and dense perception of observed interactions within a single model.

\textbf{Generative priors for dense perception.}
Marigold and Lotus demonstrate the value of generative priors for image perception~\citep{rw_ke2024marigold,rw_he2025lotus}, while Geo4D, DVD, and GenCeption extend this direction to video perception~\citep{rw_jiang2025geo4d,zhang2026dvd,rw_wang2026genception}.
Diff-2-in-1 and UniGP further connect generation and perception at the image level~\citep{rw_zheng2025diff2in1,rw_guo2026unigp}.
Building on these advances, \modelname{} jointly learns future video generation and dense perception for embodied interactions, using a shared Video DiT and a common velocity prediction space.

\textbf{Dense visual perception.}
Depth Anything 3~\citep{lin2025depthanything3}, Video Depth Anything~\citep{chen2025videodepthanything}, and VGGT-$\Omega$~\citep{vggtomega2026} estimate depth or scene geometry, while LingBot-Depth~\citep{rw_tan2026lingbotdepth} completes missing or unreliable sensor depth.
Grounding DINO~\citep{liu2024groundingdino} enables text-conditioned object localization, and SAM 2~\citep{ravi2024sam2} and SAM 3~\citep{meta2025sam3} support promptable segmentation.
For motion perception, RAFT~\citep{teed2020raft} estimates optical flow and AllTracker~\citep{harley2025alltracker} tracks dense correspondences across video frames.
RAFT-3D~\citep{teed2021raft3d}, Any4D~\citep{karhade2025any4d}, V-DPM~\citep{sucar2026vdpm}, and Track4World~\citep{lu2026track4world} further estimate 3D motion or dense spatiotemporal correspondence.
These models provide complementary estimates of scene structure and can supply pseudo-labels for multimodal learning.
\section{Method}
\label{sec:method}

\modelname{} unifies future video generation and dense perception within a single model. 
Given an initial image $I_0$ and a text prompt $c$, it synthesizes a future RGB video together with aligned embodiment segmentation, depth, and first-frame-relative 2D flow. 
Given an observed video, the same model predicts each dense modality in a single step without iterative denoising. 
We denote an RGB video by $V=\{V_t\}_{t=0}^{T}$ and its aligned embodiment segmentation, depth, and first-frame-relative 2D flow by $S=\{S_t\}_{t=0}^{T}$, $D=\{D_t\}_{t=0}^{T}$, and $F=\{F^{2\mathrm{D}}_{0\rightarrow t}\}_{t=0}^{T}$, respectively. 
These modalities characterize complementary aspects of embodied dynamics: appearance ($V$), semantics ($S$), geometry ($D$), and motion ($F$).

\begin{figure}[t]
\centering
\includegraphics[width=\linewidth]{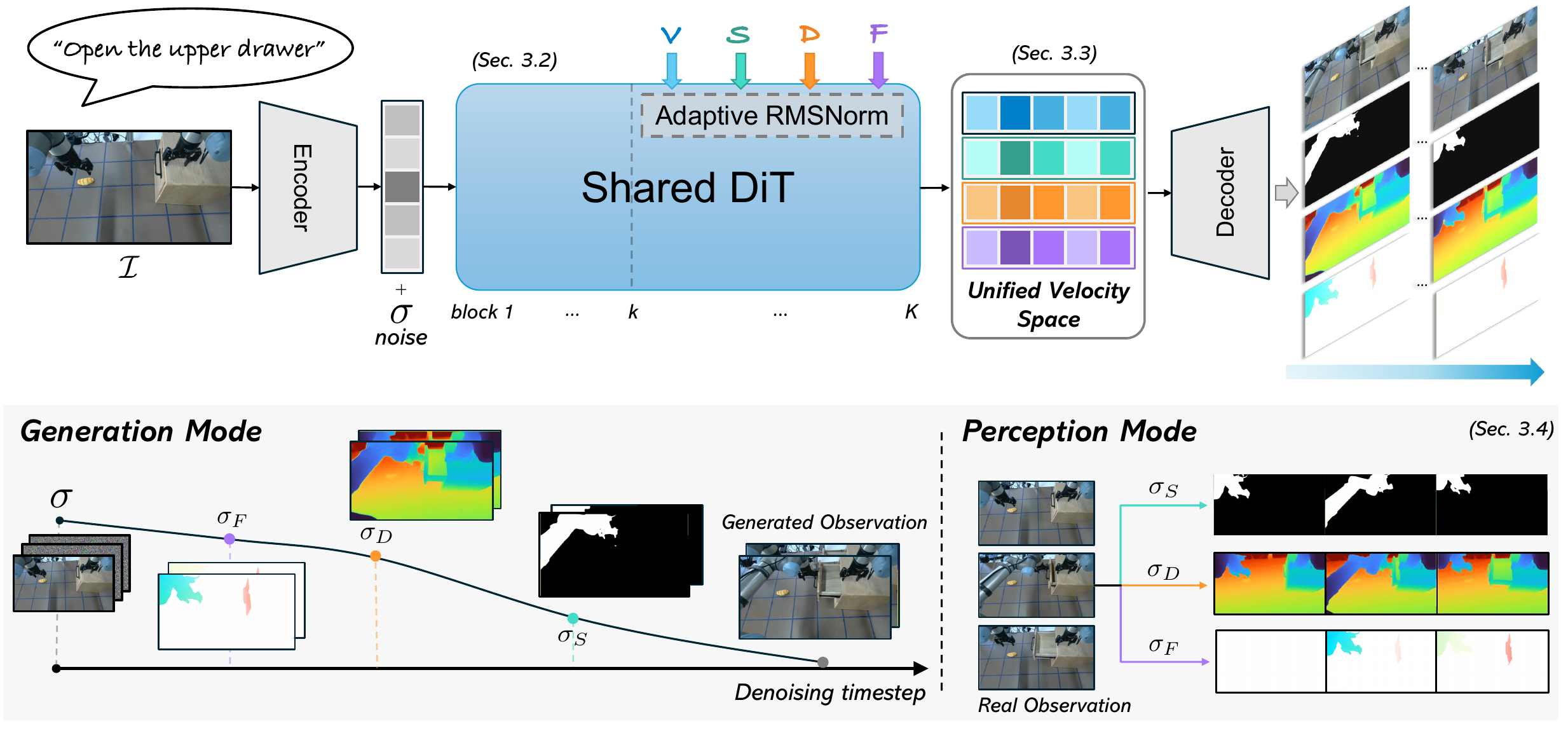}
\vspace{-6mm}
\caption{\textbf{Overview of \modelname{}.}
\textbf{Top:} A shared Video DiT processes noisy video latents and predicts velocities for RGB video ($V$), embodiment segmentation ($S$), depth ($D$), and first-frame-relative 2D flow ($F$) in a unified prediction space.
The first $k$ blocks compute task-agnostic features, while the remaining blocks apply task-adaptive RMSNorm with shared Transformer weights.
The frozen video autoencoder is shared across modalities.
\textbf{Bottom left:} Given an initial image and text prompt, video generation follows a multi-step denoising trajectory, with dense predictions read at modality-specific noise levels.
\textbf{Bottom right:} Given an observed video, each dense modality is predicted in one step from a noisy state at its selected noise level.
Both regimes share the dense readout and decoding procedure, without independent denoising trajectories for dense modalities.
}
\label{fig:pipeline}
\end{figure}

\modelname{} maps all modalities into a common image-like representation processed by the same frozen video autoencoder~\citep{wan2025wan22}. 
A shared Video DiT predicts each modality through lightweight task modulation, while generation and perception are jointly learned in a unified velocity prediction space. 
Figure~\ref{fig:pipeline} illustrates the overall pipeline.

\subsection{Unified Modality Representation}
\label{sec:method_representation}

\textbf{Everything-as-RGB Representation.}
A central challenge is to accommodate heterogeneous perceptual outputs within a model pretrained for RGB video generation. 
Rather than introducing modality-specific encoders or decoders, we confine modality-specific processing to the representation boundary and map each modality to a three-channel image-like representation. 
At training time, for each dense modality $m\in\{S,D,F\}$, an encoding mapping $\psi_m$ converts the modality data $M_m$ into an RGB-like sequence, which is then encoded by the same frozen video autoencoder used for RGB to obtain the latent target $z_m=\mathcal{E}(\psi_m(M_m))$.
At inference time, the shared decoder $\mathcal{D}$ reconstructs an image-like prediction, which is converted back to the corresponding modality space through a deterministic decoding mapping $\rho_m$.
Together, $\psi_m$ and $\rho_m$ define the representation interface and introduce no learnable modality-specific representation modules. 
Detailed definitions of the modality-specific transformations $\psi_m$ and $\rho_m$ are provided in Appendix~\ref{app:encodings}.

\textbf{Recovering 3D Scene Flow.}
The predicted depth and first-frame-relative 2D flow $F^{2\mathrm{D}}_{0\rightarrow t}$ can be combined to recover the 3D scene flow $F^{3\mathrm{D}}_{0\rightarrow t}$. 
For a pixel $p$ in the initial frame, its corresponding location at time $t$ is $p_t = p + F^{2\mathrm{D}}_{0\rightarrow t}(p)$. Assuming a static camera with constant intrinsic matrix $K$, the world coordinate system is aligned with the camera coordinate frame, simplifying inter-frame camera motion to the identity transformation. 
By back-projecting the corresponding pixels into the unified 3D camera coordinate frame, the 3D scene flow is obtained as:
\begin{equation}
P_0 = D_0(p) \cdot K^{-1} \tilde{p}, \qquad P_t = D_t(p_t) \cdot K^{-1} \tilde{p}_t, \qquad F^{3\mathrm{D}}_{0\rightarrow t}(p) = P_t - P_0,
\label{eq:scene_flow_lift}
\end{equation}
where $\tilde{p} = [p^\top, 1]^\top$ and $\tilde{p}_t = [p_t^\top, 1]^\top$ denote homogeneous coordinates of $p$ and $p_t$.

\subsection{Unified Model Architecture}
\label{sec:method_architecture}

\textbf{Shared Video DiT Backbone.}
All four tasks are predicted by a single, fully shared Video DiT.
The model retains the standard video-generation input formulation, while embodiment segmentation, depth,  and 2D flow are predicted as additional outputs rather than provided as input modalities. 
Let the backbone contain $L$ Transformer blocks.
The input is processed once through the first $k$ blocks to obtain task-agnostic spatiotemporal features,
\begin{equation}
h_k=\mathcal{B}_{1:k}(x_\sigma,\sigma,c),
\label{eq:shared_trunk}
\end{equation}
where $x_\sigma$ denotes the noisy video latent at noise level $\sigma$, $c$ denotes the text condition, and $\mathcal{B}_{1:k}$ denotes the first $k$ Transformer blocks.
The resulting activations $h_k$ are reused across all tasks. 
The remaining $L-k$ blocks and output projection process the task-conditioned streams in parallel with shared parameters, while task dependence is introduced solely through the adaptive modulation described below.
The choice of $k$ controls the balance between shared computation and task-conditioned processing.
We use $k=16$ by default and study the effect of $k$ in Sec.~\ref{sec:exp_ablation}.

\textbf{Task Modulation via Adaptive RMSNorm.}
The pretrained Video DiT used in our model employs RMSNorm~\citep{zhang2019root}, which we extend with lightweight task-adaptive modulation.
For a hidden representation $h\in\mathbb{R}^{d}$, the pretrained RMSNorm is written as
\begin{equation}
\operatorname{RMSNorm}(h)
=
\gamma\odot
\frac{h}{\sqrt{d^{-1}\sum_i h_i^2+\varepsilon}},
\end{equation}
where $\gamma$ is the learned scaling parameter inherited from the pretrained model. 
For each task-conditioned block $\ell$ and task $\tau\in\{V,S,D,F\}$, a learned modulation table directly provides the corresponding scale, shift, and residual-gate deltas,
\begin{equation}
(\Delta s_{\ell,\tau},\Delta b_{\ell,\tau},\Delta g_{\ell,\tau})
=
\mathcal{T}(\ell,\tau).
\end{equation}
These deltas modulate the normalized features and residual update as
\begin{equation}
\widetilde h_{\ell,\tau}
=
(1+\Delta s_{\ell,\tau})\odot\operatorname{RMSNorm}(h)
+\Delta b_{\ell,\tau},
\qquad
h'
=
h+(1+\Delta g_{\ell,\tau})\odot\mathcal{F}(\widetilde h_{\ell,\tau}),
\label{eq:adaptive_rmsnorm}
\end{equation}
where $\mathcal{F}$ denotes the corresponding attention or feed-forward transformation.
The attention and feed-forward sub-blocks use separate modulation tables, while all Transformer parameters remain shared across tasks.
All modulation parameters are initialized to zero, such that the scale and residual-gate factors start at one and the shifts at zero, exactly preserving the pretrained computation at initialization.
Task-specific behavior is then learned through these lightweight modulations during joint training.
The RMSNorm preceding the final output projection is modulated in the same manner using task-specific scale and shift, without a residual gate. 
Consequently, the video autoencoder, Transformer parameters, and output projection are fully shared across modalities, with task specialization introduced only through lightweight task-adaptive modulation.

\subsection{Unified Prediction Space}
\label{sec:method_velocity}

\begin{wrapfigure}{r}{0.5\textwidth}
    \vspace{-10pt}
    \centering
    \includegraphics[width=\linewidth]{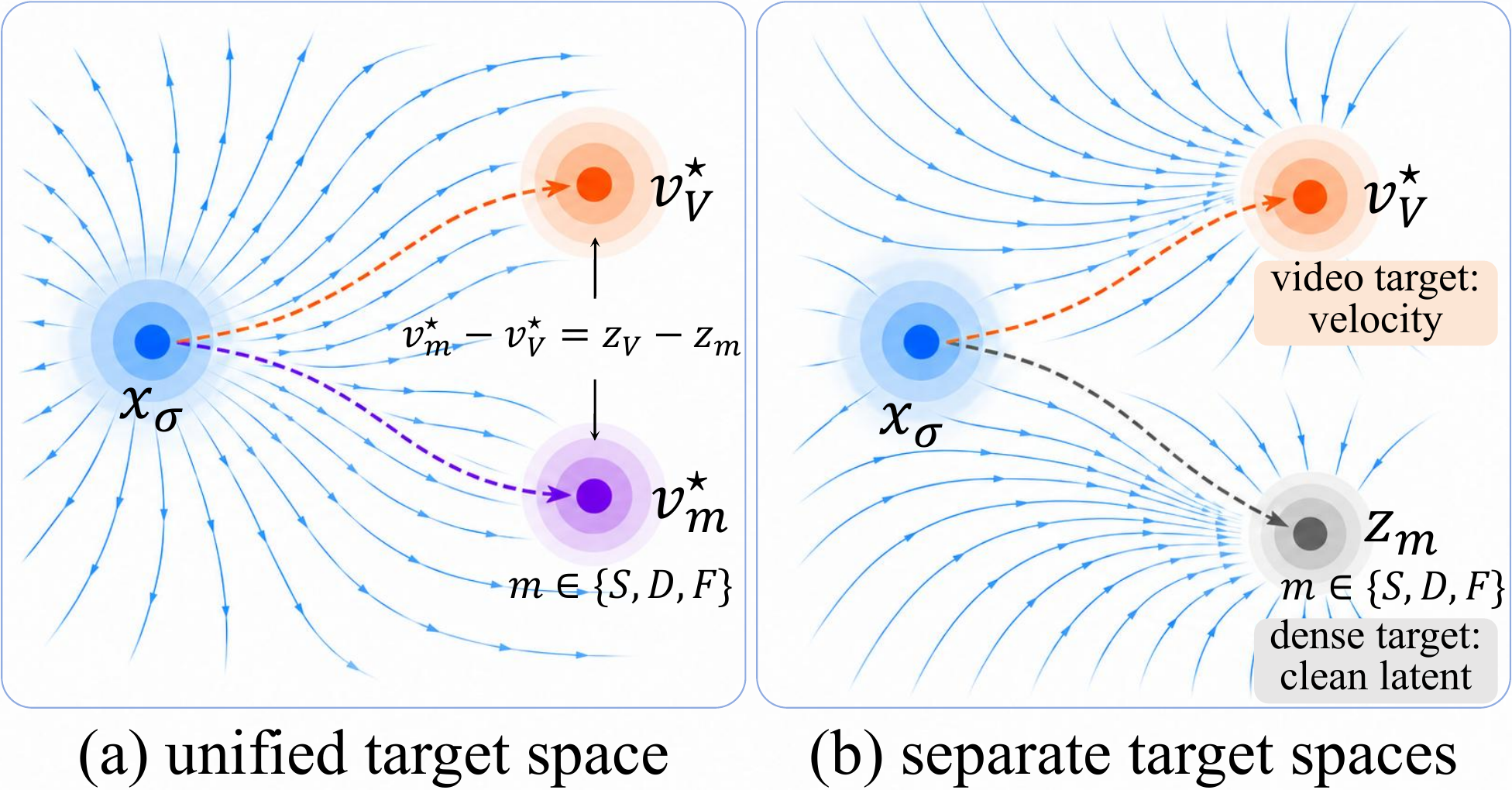}
    \caption{Comparison of prediction target spaces.}
    \label{fig:vel_space}
\end{wrapfigure}

\textbf{Unified Velocity Prediction.}
Beyond sharing the modality representation and model architecture, \modelname{} unifies the prediction parameterization of generation and dense perception.
A natural alternative is to directly regress the clean dense latents while retaining velocity prediction for video (Figure~\ref{fig:vel_space}(b)).
This would require the shared model to serve two different target parameterizations: video velocities and clean perceptual latents.
We instead retain the pretrained video generator's velocity parameterization across all tasks (Figure~\ref{fig:vel_space}(a)).

Let $z_V$ denote the clean video latent and $z_m$ the clean latent of dense modality $m\in\{S,D,F\}$.
For a noise level $\sigma\in[0,1]$ and a Gaussian noise sample $\epsilon\sim\mathcal{N}(0,I)$, we follow the linear interpolation path used in rectified flow and flow matching~\citep{liu2023rectifiedflow,lipman2023flowmatching} and define
\begin{equation}
x_\sigma = (1-\sigma)z_V+\sigma\epsilon, \qquad
v_V^\star = \epsilon-z_V, \qquad
v_m^\star = \epsilon-z_m,
\quad m\in\{S,D,F\}.
\label{eq:unified_velocity}
\end{equation}
All task streams use the same noisy video input $x_\sigma$ at the same noise level $\sigma$, and all velocity targets are constructed using the same noise sample $\epsilon$.
The dense latents serve as supervision targets rather than additional input modalities.
The interpolation above applies only to the frames to be predicted; the initial frame remains clean as image conditioning and is excluded from all prediction losses.

\textbf{Cross-Modal Residual Interpretation.}
The shared noise establishes a direct relation between the video and dense velocity targets:
\begin{equation}
v_m^\star-v_V^\star
=
(\epsilon-z_m)-(\epsilon-z_V)
=
z_V-z_m.
\label{eq:cross_modal_residual}
\end{equation}
The noise cancels exactly in their difference, leaving the cross-modal residual between the clean video and dense latents.
Thus, relative to the video target, each dense target differs by a noise-free cross-modal residual while retaining the same velocity parameterization.
This complements the zero-initialized task modulation in Sec.~\ref{sec:method_architecture}, under which all task streams initially inherit the pretrained video generator.
The formulation therefore provides a common starting point for task-specific adaptation without switching the dense streams to clean-latent prediction.
The dense modalities do not require independent denoising processes. 
Their reconstruction from the jointly predicted velocities is described in the next subsection.

\textbf{Joint Training Objective.}
Let $\hat v_\tau$ denote the model's velocity prediction for task $\tau\in\{V,S,D,F\}$.
Each task is supervised by the velocity regression loss
$\mathcal{L}^{\mathrm{vel}}_\tau=\|\hat v_\tau-v_\tau^\star\|_2^2$.
For the dense modalities, we additionally apply a spatial-gradient matching loss~\citep{eigen2015predicting} to supervise local spatial variation. 
The joint training objective is
\begin{equation}
\mathcal{L}
=
\sum_{\tau\in\{V,S,D,F\}}
w_\tau\lambda_\tau(n)\mathcal{L}_\tau,\qquad
\mathcal{L}_V
=
\mathcal{L}^{\mathrm{vel}}_V,\qquad
\mathcal{L}_m
=
\mathcal{L}^{\mathrm{vel}}_m
+ \tfrac{1}{2} 
\|\nabla\hat v_m-\nabla v_m^\star\|_2^2.
\label{eq:training_objective}
\end{equation} 
Here, $w_\tau$ denotes the task weight, and $n$ denotes the training step. 
The operator $\nabla$ computes discrete spatial gradients along the height and width dimensions of the latent velocity predictions and targets. 
We use $w_V=1$ and $w_S=w_D=w_F=0.25$ by default. 
We set $\lambda_V(n)=1$, while the dense-task warmup factors $\lambda_m(n)$ gradually increase from zero to one during early training to reduce disruption to the pretrained generation capability.

\subsection{Inference for Generation and Perception}
\label{sec:method_inference}

Building on the unified velocity formulation, \modelname{} supports two complementary inference regimes with shared parameters and a common dense readout. 
In generation, dense predictions are read from intermediate states along the video denoising trajectory. 
In perception, noisy states are constructed directly from an observed video. 
Figure~\ref{fig:pipeline} illustrates the two regimes.

\textbf{Multi-Step Video Generation with Dense Readouts.} 
Given an initial image $I_0$ and a text prompt $c$, \modelname{} generates a future RGB video through $K$ denoising steps,
$x_{\sigma_K}\rightarrow x_{\sigma_{K-1}}\rightarrow\cdots\rightarrow x_{\sigma_0}$,
with decreasing noise levels.  
For each dense modality $m\in\{S,D,F\}$, we empirically select a readout noise level $\sigma_m^\star$, which is used in both inference regimes.
The choice of $\sigma_m^\star$ is studied in Sec.~\ref{sec:exp_ablation}.
During generation, the corresponding dense task stream is run only when the video denoising trajectory reaches this readout state.

At $x_{\sigma_m^\star}$, the video and dense task streams produce $\hat v_V$ and $\hat v_m$ from the same input.
The video prediction yields the noise estimate
$\hat\epsilon=x_{\sigma_m^\star}+(1-\sigma_m^\star)\hat v_V$,
and the dense latent is reconstructed as
\begin{equation}
\begin{aligned}
\hat z_m
=
\hat\epsilon-\hat v_m
&=
x_{\sigma_m^\star}
+
(1-\sigma_m^\star)\hat v_V
-
\hat v_m\\
&=
\hat z_V-(\hat v_m-\hat v_V),
\end{aligned}
\label{eq:dense_readout}
\end{equation}
where $\hat z_V=x_{\sigma_m^\star}-\sigma_m^\star\hat v_V$ is the model's clean video latent estimate at the readout state.
The difference $\hat v_m-\hat v_V$ estimates the cross-modal residual $z_V-z_m$ in Eq.~(\ref{eq:cross_modal_residual}). 
Thus, the dense latent is reconstructed relative to the model's video estimate.
The corresponding prediction is obtained as
$\hat M_m=\rho_m(\mathcal{D}(\hat z_m))$,
using the shared video decoder $\mathcal{D}$ and the modality decoding mapping $\rho_m$.
Dense predictions are therefore obtained without independent denoising trajectories.

\textbf{Single-Step Dense Perception.}
Given an observed video, we encode it as $z_V$.
For a requested modality $m$, we directly construct the noisy state at its selected noise level,
\begin{equation}
x_{\sigma_m^\star}
=
(1-\sigma_m^\star)z_V
+
\sigma_m^\star\epsilon,
\qquad
\epsilon\sim\mathcal{N}(0,I).
\end{equation}
A single forward pass at this state produces both the video velocity $\hat v_V$ and the dense velocity $\hat v_m$.
We then apply the readout in Eq.~(\ref{eq:dense_readout}) and the same decoding procedure to obtain $\hat M_m$.
Each requested modality is therefore predicted from its selected noisy state without iterative denoising.

\newcommand{\gsFirst}[1]{{\setlength{\fboxsep}{1pt}\colorbox[RGB]{180,225,245}{#1}}}
\newcommand{\gsSecond}[1]{{\setlength{\fboxsep}{1pt}\colorbox[RGB]{210,235,248}{#1}}}

\section{Experiments}
\label{sec:exp}

We evaluate \modelname{} as both a video generator and a dense perception model. The experiments examine \textit{video quality}, the accuracy of \textit{depth}, \textit{embodiment segmentation}, and \textit{scene flow} prediction in both \textbf{generation mode} and \textbf{perception mode}, and generalization to unseen manipulation tasks and a simulated environment.  
Our ablation studies address two questions: (1) How many task-agnostic blocks should be used? (2) Which readout noise level is best suited to each dense task?

\subsection{Datasets and Metrics}
\label{sec:exp_datasets}

We train \modelname{} on 221,044 episodes across four embodiments from RoboMIND~2.0~\citep{hou2025robomind2} and our egocentric human dataset \emph{EgoTable}. Additionally, we evaluate task-level generalization on out-of-distribution (OOD) tasks and assess cross-domain transfer to simulation via RoboTwin~2.0~\citep{chen2025robotwin2}. Dataset composition, split protocols, and dense labels are detailed in Appendix~\ref{app:datasets} and~\ref{app:pseudolabels}.

\textbf{For video generation,} 
we evaluate 49-frame text-image-to-video generation using FVD~\citep{unterthiner2018fvd}, FID~\citep{heusel2017fid}, PSNR, and LPIPS~\citep{zhang2018lpips}, conditioned on the initial frame and caption. 
\textbf{For embodiment segmentation,} 
we report intersection over union (IoU, \%) for masks, with  robot arm/gripper or human hand as foreground.
\textbf{For depth,} 
we report root mean squared error (RMSE, centimeters), absolute relative error (AbsRel, \%), and threshold accuracy ($\delta_1$, \%) over valid pixels. 
\textbf{For scene flow,} 
we report first-frame-relative 2D endpoint error (EPE) and 3D average/final errors (ADE/FDE) in absolute and displacement forms. For generated 2D flow, we additionally report nEPE and $\mathrm{AAE}_{\mathrm{top}\,1\%}$. Detailed definitions are in Appendix~\ref{app:flow_metrics}.

\subsection{Comparisons with Baselines}
\label{sec:exp_comparisons}

\subsubsection{Generation Mode}
\textbf{Baselines.}
We compare generated RGB video with Wan~2.2~\citep{wan2025wan22}, 
RynnWorld4D~\citep{zhao2026rynnworld4d4dembodiedworld}, and Cosmos3-Nano~\citep{nvidia2026cosmos3}.
RynnWorld4D builds on Wan~2.2, while \modelname{} builds on Cosmos3-Nano.
RynnWorld4D jointly generates RGB video, depth, and optical flow, but not embodiment segmentation; we therefore compare it with \modelname{} on TI2V generation, depth, and flow.

\begin{table}[t]
\centering
\captionsetup{position=top,skip=4pt}
\caption{\textbf{Generation mode} on RoboMIND \textit{task-OOD} and \textit{OOD} RoboTwin set. Depth and flow are scored with VGGT-$\Omega$ and AllTracker, respectively. AbsRel and $\delta_1$ are in \%; $\mathrm{AAE}_{\mathrm{top}\,1\%}$ is in degrees; EPE is in pixels.}
\label{tab:gs_video}
\footnotesize
\setlength{\tabcolsep}{0.25pt}
\renewcommand{\arraystretch}{1.02}
\begin{tabular*}{\linewidth}{@{\extracolsep{\fill}}lrrrrrrrrr@{}}
\toprule
& \multicolumn{4}{c}{Video} & \multicolumn{2}{c}{Depth} & \multicolumn{3}{c}{2D Flow} \\
\cmidrule(lr){2-5}\cmidrule(lr){6-7}\cmidrule(lr){8-10}
\textbf{Method} & FVD $\downarrow$ & FID $\downarrow$ & PSNR $\uparrow$ & LPIPS $\downarrow$ & AbsRel $\downarrow$ & $\delta_1\uparrow$ & nEPE $\downarrow$ & $\mathrm{AAE}_{\mathrm{top}\,1\%}\downarrow$ & EPE $\downarrow$ \\
\midrule
\multicolumn{10}{@{}l}{\textit{RoboMIND 2.0}} \\
Wan2.2-I2V-A14B & 69.37 & 50.01 & 17.13 & 0.3083 & -- & -- & -- & -- & -- \\
RynnWorld4D & 73.45 & 47.90 & \textbf{18.07} & 0.2661 & 8.53 & 92.55 & 1.133 & 45.14 & \textbf{4.936} \\
Cosmos3-Nano & 44.97 & 42.09 & 16.90 & 0.3002 & -- & -- & -- & -- & -- \\
\textbf{\modelname{}} & \textbf{14.55} & \textbf{16.12} & 18.01 & \textbf{0.2296} & \textbf{7.88} & \textbf{93.17} & \textbf{0.427} & \textbf{23.10} & 7.089 \\
\midrule
\multicolumn{10}{@{}l}{\textit{RoboTwin~2.0}} \\
RynnWorld4D & 71.18 & 67.65 & 16.53 & 0.3839 & \textbf{4.86} & 95.69 & 1.225 & 47.49 & \textbf{7.11} \\
\textbf{\modelname{}} & \textbf{61.14} & \textbf{46.28} & \textbf{16.71} & \textbf{0.2927} & 4.98 & \textbf{97.25} & \textbf{0.892} & \textbf{26.78} & 16.45 \\
\bottomrule
\end{tabular*}
\end{table}

\textbf{Results and discussion.}
Table~\ref{tab:gs_video} reports the quantitative comparison, with qualitative examples shown in Figure~\ref{fig:gs_comparison}.
On RoboMIND, \modelname{} achieves the lowest FVD, FID, and LPIPS among the four generators, with PSNR comparable to RynnWorld4D. Its generated depth also improves both AbsRel and $\delta_1$. For 2D flow, \modelname{} has lower nEPE and $\mathrm{AAE}_{\mathrm{top}\,1\%}$, whereas RynnWorld4D has lower raw EPE, indicating how EPE alone can favor low-motion generations.
On RoboTwin~2.0, \modelname{} improves all four video metrics and depth $\delta_1$.
The flow trend is consistent across datasets: \modelname{} has lower nEPE and top-motion angular error, while RynnWorld4D has lower raw EPE.

% --------------------------------
\subsubsection{Perception Mode}

\begin{figure}[t]
\centering
\includegraphics[width=\linewidth]{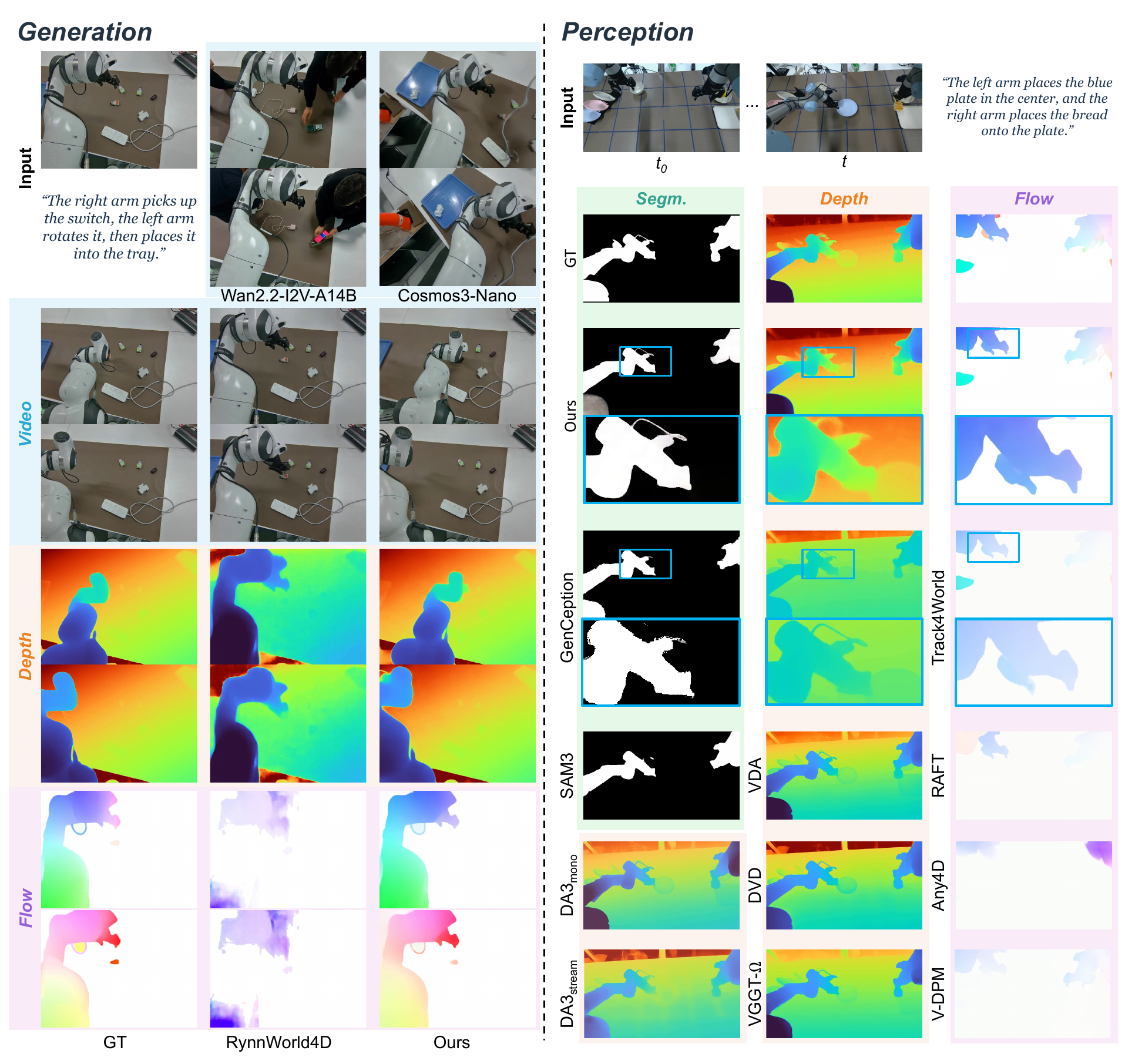}
\vspace{-5mm}
\caption{\textbf{Qualitative comparison in generation and perception modes.}
\textbf{Left (generation):} Given an initial frame and action prompt, the top panels show video examples from Wan~2.2 and Cosmos3-Nano; below, two time frames of video, depth, and 2D flow compare the reference, RynnWorld4D, and \modelname{}.
\textbf{Right (perception):} Given observed RGB frames, we compare \modelname{} with reference maps and task-specific baselines for embodiment segmentation, depth, and 2D flow. Blue boxes mark zoomed-in regions.}
\label{fig:gs_comparison}
\end{figure}

\begin{table}[t]
\centering
\captionsetup{position=top,skip=4pt}
\caption{\textbf{Perception mode} on RoboMIND \textit{task-OOD} set. RAFT$^{\dagger}$ and RAFT$^{\ddagger}$ share the 2D EPE, and their 3D scores use ground-truth and our predicted depth, respectively. Ranking is per metric: \gsFirst{1st}, \gsSecond{2nd}.}
\label{tab:gs_perception}
\captionsetup[subtable]{position=top,skip=3pt}
\footnotesize
\begin{subtable}[t]{0.18\linewidth}
\centering
\caption{\textbf{Segmentation}}
\label{tab:gs_segmentation}
\setlength{\tabcolsep}{0.25pt}
\begin{tabular*}{\linewidth}{@{\extracolsep{\fill}}lr@{}}
\toprule
\textbf{Method} & IoU $\uparrow$ \\
\midrule
SAM3 & \gsSecond{73.5} \\
GenCeption & 67.5 \\
\midrule
\textbf{\modelname{}} & \gsFirst{88.7} \\
\bottomrule
\end{tabular*}
\end{subtable}\hfill
\begin{subtable}[t]{0.34\linewidth}
\centering
\caption{\textbf{Depth}}
\label{tab:gs_depth}
\setlength{\tabcolsep}{0.25pt}
\begin{tabular*}{\linewidth}{@{\extracolsep{\fill}}lrrr@{}}
\toprule
\textbf{Method} & RMSE $\downarrow$ & AbsRel $\downarrow$ & $\delta_1\uparrow$ \\
\midrule
VDA & 13.12 & 7.4 & 94.0 \\
DA3$_\text{mono}$ & 15.65 & 13.2 & 81.7 \\
DA3$_\text{stream}$ & 16.17 & 13.2 & 85.3 \\
DVD & 15.91 & 7.6 & 92.9 \\
VGGT-$\Omega$ & \gsSecond{13.11} & \gsSecond{5.9} & \gsSecond{94.5} \\
GenCeption & 15.89 & 8.5 & 90.7 \\
\midrule
\textbf{\modelname{}} & \gsFirst{4.76} & \gsFirst{2.4} & \gsFirst{99.1} \\
\bottomrule
\end{tabular*}
\end{subtable}\hfill
\begin{subtable}[t]{0.44\linewidth}
\centering
\caption{\textbf{Flow}}
\label{tab:gs_flow}
\setlength{\tabcolsep}{0.25pt}
\begin{tabular*}{\linewidth}{@{\extracolsep{\fill}}lccc@{}}
\toprule
\textbf{Method} & EPE $\downarrow$ & ADE$_{\mathrm{abs/disp}}\downarrow$ & FDE$_{\mathrm{abs/disp}}\downarrow$ \\
\midrule
RAFT$^{\dagger}$ & \multirow{2}{*}{11.2} & \gsFirst{7.04}/6.88 & \gsFirst{7.49}/\gsSecond{7.40} \\
RAFT$^{\ddagger}$ & & 9.63/7.68 & 9.95/7.92 \\
Any4D & 18.4 & 15.26/\gsSecond{6.72} & 16.93/7.97 \\
V-DPM & 25.2 & 24.17/\gsFirst{4.88} & 26.24/\gsFirst{7.14} \\
Track4World & \gsSecond{6.8} & 20.31/6.92 & 23.99/9.06 \\
\midrule
\textbf{\modelname{}} & \gsFirst{5.2} & \gsSecond{7.39}/7.68 & \gsSecond{9.51}/9.87 \\
\bottomrule
\end{tabular*}
\end{subtable}
\end{table}

With ground-truth RGB video as input, we compare \modelname{} against task-specific foundation models on embodiment segmentation, depth, and flow, as shown in Figure~\ref{fig:gs_comparison}.

\textbf{Embodiment segmentation.}
We compare with text-prompted SAM3~\citep{meta2025sam3} and GenCeption~\citep{rw_wang2026genception}, using the same embodiment prompt for both baselines.
Table~\ref{tab:gs_segmentation} shows that \modelname{} achieves the highest IoU, indicating that its shared representation can also capture embodiment semantics effectively.

\textbf{Depth estimation.}
We compare with DA3~\citep{lin2025depthanything3} in monocular and streaming configurations, VGGT-$\Omega$~\citep{vggtomega2026}, DVD~\citep{zhang2026dvd}, VDA~\citep{chen2025videodepthanything}, and GenCeption~\citep{rw_wang2026genception}. 
As shown in Table~\ref{tab:gs_depth}, \modelname{} performs best across RMSE, AbsRel, and $\delta_1$, while VGGT-$\Omega$ is the strongest baseline across these metrics. This suggests that the shared video backbone also supports accurate geometric perception.

\textbf{Flow estimation.}
For 2D flow, we compare with RAFT~\citep{teed2020raft}, Any4D~\citep{karhade2025any4d}, V-DPM~\citep{sucar2026vdpm}, and Track4World~\citep{lu2026track4world}. For 3D scene flow, we compare with RAFT-3D~\citep{teed2021raft3d}, Any4D, V-DPM, and Track4World.
Table~\ref{tab:gs_flow} shows \modelname{} achieves the lowest 2D EPE. For 3D flow, it ranks behind only RAFT-3D with ground-truth depth and outperforms all methods without ground-truth depth.
Note that we use absolute trajectory errors as the primary 3D metric, since displacement errors can partially cancel depth discontinuities and alter the ranking.
Further analysis is provided in Appendix Figure~\ref{fig:tracking_3d}. 

\begin{figure}[t]
\centering
\includegraphics[width=\linewidth]{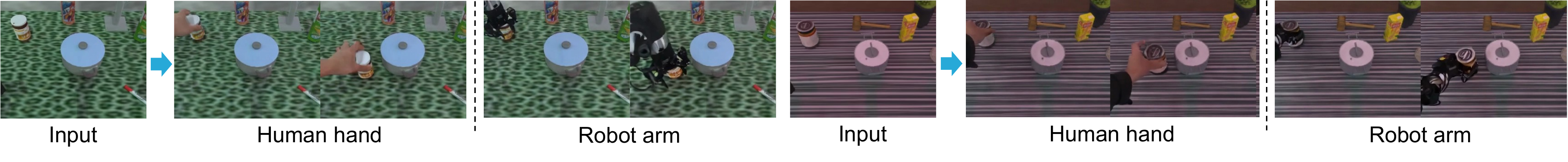}
\vspace{-0.7cm}
\caption{\textbf{Embodiment-conditioned generation on RoboTwin~2.0.} Given the same initial frame (left), \modelname{} generates a human-hand continuation (middle) or a robot-arm continuation (right) under the corresponding embodiment condition.}
\label{fig:gs_robotwin}
\vspace{-0.2cm}
\end{figure}

\textbf{Generalization across embodiments and to simulation.}
Figure~\ref{fig:gs_robotwin} qualitatively demonstrates GenSight's embodiment controllability on RoboTwin 2.0. Given the same initial frame, it successfully synthesizes continuation videos for either a human hand or a robot arm in simulated scenes.   
Additional zero-shot generalization results are shown in Appendix Figure~\ref{fig:robotwin_2.0}.

\vspace{-1mm}
\subsection{Ablations}
\label{sec:exp_ablation}
We examine the task-agnostic prefix length and dense-readout noise levels on held-out UR5 tasks. 
Experimental settings and detailed per-metric results are provided in Appendix~\ref{app:ablation_details}.

\begin{figure*}[t]
\centering
\includegraphics[width=\linewidth]{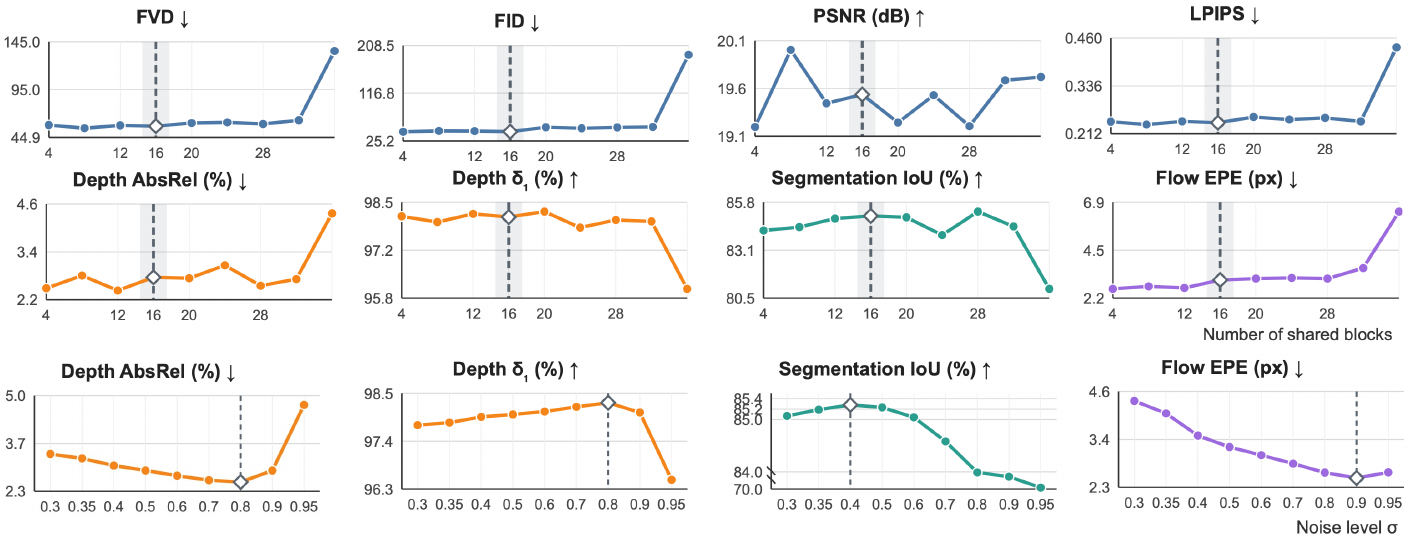}
\vspace{-0.7cm}
\caption{\textbf{Ablation studies on UR5.} 
\textbf{Top two rows:} Video generation and dense perception metrics across task-agnostic prefix lengths $k$; the dashed line marks our selected configuration. 
\textbf{Bottom row:} Evaluation of a fixed checkpoint across readout noise levels $\sigma$, where markers indicate optima.}
\label{fig:ablation_studies}
\end{figure*}

\textbf{Number of task-agnostic blocks.}
As shown in Figure~\ref{fig:ablation_studies} (top two rows), the optimal value of $k$ varies across metrics. 
A long task-agnostic prefix degrades both video generation and dense perception, whereas an intermediate prefix length provides a better overall balance. 
These results support preserving task-adaptive modulation in the later shared blocks.

\textbf{Noise level for dense prediction.} 
Figure~\ref{fig:ablation_studies} (bottom row) illustrates that different dense modalities favor distinct readout noise levels. 
Depth performs best around $\sigma=0.8$, while segmentation prefers a lower noise level of $\sigma=0.4$, and flow continues to improve up to $\sigma=0.9$. 
These results motivate using separate readout noise levels for different dense tasks.

\section{Conclusion}
\label{sec:conclusion}

We presented \modelname{}, a unified embodied world model that brings future video generation and dense perception into a single pretrained Video DiT. 
\modelname{} unifies heterogeneous modalities through a common spatiotemporal representation, a shared model architecture with lightweight task adaptation, and a unified velocity prediction space. 
This formulation enables multi-step future video generation with aligned dense readouts and single-step dense perception from observed videos. 
Experiments across diverse embodiments, human-hand interactions, unseen tasks, and simulated environments demonstrate strong performance in both regimes and validate the key design choices. 
Overall, our results show that seeing and imagining can be jointly modeled within a single embodied world model, providing a unified foundation for understanding embodied dynamics.

\textbf{Limitations and Future Work.}
\modelname{} relies partly on sensor completion and off-the-shelf estimators for dense supervision, which may introduce annotation noise and bias. 
Moreover, the current model focuses on world modeling and does not yet close the loop with action prediction or control. 
Future work will explore more reliable dense supervision and tighter integration of generation, perception, and action for closed-loop embodied interaction.

% \subsubsection*{Acknowledgments}
% Use unnumbered third level headings for the acknowledgments. All
% acknowledgments, including those to funding agencies, go at the end of the paper.

\bibliography{references}
\bibliographystyle{rbs}

\clearpage
\appendix
\setcounter{figure}{0}
\setcounter{table}{0}
\renewcommand{\thefigure}{\Alph{figure}}
\renewcommand{\thetable}{\Alph{table}}
\renewcommand{\theHfigure}{appendix.\Alph{figure}}
\renewcommand{\theHtable}{appendix.\Alph{table}}

\begin{center}
    {\Large\bf Supplementary Material for \\[2pt]
    ``Seeing and Imagining as One: Unified Video Generation and 4D Perception for Embodied World Modeling''}
\end{center}
\vspace{1em}

\section{Dense modality encoding and decoding}
\label{app:encodings}
\subsection*{1. Depth Representation}
We encode metric depth in the logarithmic domain to retain resolution at close range while compressing large distances. The normalization bounds are determined once from non-sky depth values in the training split: $d_{1}$ and $d_{99}$ denote the 1st and 99th percentiles, respectively. We use these fixed bounds for every training and inference sample, avoiding sample-dependent scaling and the influence of extreme depth values. The normalized depth is then represented by a three-channel RGB thermometer code, which provides finer quantization than a single 8-bit channel.

\paragraph{Depth Normalization}
For a metric depth map $D$, let $y=\ln D$, $y_{\min}=\ln d_{1}$, and $y_{\max}=\ln d_{99}$. We normalize and clip log depth to the fixed training-set interval:
\begin{equation}
\hat{D} = \operatorname{clip}\left(\frac{\ln D-y_{\min}}{y_{\max}-y_{\min}},\,0,\,1\right).
\end{equation}
The same $y_{\min}$ and $y_{\max}$ are used for all samples during both training and testing.

\paragraph{Depth Encoding}
We quantize $\hat{D}\in[0,1]$ as $S=\lfloor 767\hat{D}\rceil\in\{0,\ldots,767\}$ and map it to an RGB thermometer code $\mathbf{C}=[R,G,B]^\top$ by clamping each channel to $[0,255]$:
\begin{equation}
\mathbf{C}(\hat{D}) = \min\left(\max\left( \begin{bmatrix} S \\ S - 256 \\ S - 512 \end{bmatrix}, \begin{bmatrix} 0 \\ 0 \\ 0 \end{bmatrix} \right), \begin{bmatrix} 255 \\ 255 \\ 255 \end{bmatrix} \right).
\end{equation}
Here, the minimum and maximum are applied elementwise.

\paragraph{Depth Decoding}
Given $\mathbf{C}$, we sum the channel values and invert the log-domain normalization:
\begin{equation}
S = R+G+B, \qquad \hat{D}=\frac{S}{767},
\end{equation}
\begin{equation}
D=\exp\left(y_{\min}+\hat{D}(y_{\max}-y_{\min})\right).
\end{equation}
Decoding uses the same fixed training-set bounds and does not access test-set depth statistics.

\subsection*{2. Segmentation Representation}
We represent robot arms, grippers, and human hands with three-channel encodings of binary masks. A threshold applied after decoding recovers binary segmentation masks.

\paragraph{Segmentation Encoding}
Given a binary mask $M \in \{0, 1\}^{H \times W}$, we replicate it across three RGB channels to form $\mathbf{M} \in \{0, 255\}^{3 \times H \times W}$:
\begin{equation}
\mathbf{M} = 255 \cdot \begin{bmatrix} M \\ M \\ M \end{bmatrix}
\end{equation}

\paragraph{Segmentation Decoding}
Given the reconstructed continuous values $\mathbf{\hat{M}} \in [0, 255]^{3 \times H \times W}$, the binary mask $\hat{M}$ is recovered by channel-averaging and applying a threshold $\tau = 127$:
\begin{equation}
\hat{M} = \mathbb{I}\left( \frac{1}{3} \sum_{c \in \{R,G,B\}} \mathbf{\hat{M}}_c > 127 \right)
\end{equation}

\subsection*{3. Flow Representation}
We map the two-channel displacement field to an HSV color wheel and convert it to an RGB representation compatible with the frozen visual tokenizer.

\paragraph{Flow Normalization}
Let $F^{\mathrm{grid}}\in\mathbb{R}^{T\times g_h\times g_w\times 2}$ be the raw first-frame-anchored flow, expressed in pixels of a flow grid of width $g_w$. We first rescale it onto a common canvas of width $W=1280$ using a single scalar, which leaves the displacement direction invariant:
\begin{equation}
F = \frac{W}{g_w}\, F^{\mathrm{grid}}
\end{equation}
We suppress low-magnitude trajectories using a per-pixel temporal gate: a pixel $p$ is retained only if its maximum displacement over the clip exceeds the threshold $\tau=100$; otherwise, its flow is set to zero in every frame:
\begin{equation}
K(p) = \mathbb{I}\!\left( \max_{t} \left\| F_t(p) \right\|_2 \;\ge\; 100 \right),\qquad F_t(p) \leftarrow K(p)\, F_t(p)
\end{equation}

\paragraph{Flow Encoding}
Write $\tilde{F}=(u,v)$ for the normalized flow at a pixel, expressed in polar coordinates:
\begin{equation}
\theta = \operatorname{atan2}(v, u) \in [-\pi, \pi),\qquad m = \sqrt{u^{2} + v^{2}}
\end{equation}
Direction is mapped to hue, magnitude to saturation, and value is held at its maximum. Using the 8-bit convention ($H\in\{0,\dots,179\}$):
\begin{equation}
H = \left\lfloor 179 \cdot \frac{\theta + \pi}{2\pi} \right\rfloor,\qquad S = \left\lfloor 255 \cdot \min\!\left( \frac{m}{\sigma},\, 1 \right) \right\rfloor,\qquad V = 255
\end{equation}
where $\sigma > 0$ is a single dataset-wide magnitude scale estimated as the 99th percentile of moving pixel magnitudes. The final RGB code is obtained via the standard conversion:
\begin{equation}
\mathbf{C} = \mathcal{H}^{-1}\!\left( H,\, S,\, V \right) \in \{0,\dots,255\}^{3}
\end{equation}
Zero motion and invalid pixels are both assigned the neutral white code ($S=0, V=255$) to prevent injecting spurious hue.

\paragraph{Flow Decoding}
Given a reconstructed RGB code $\hat{\mathbf{C}}$, the optical flow direction and magnitude are recovered by inverting the color wheel:
\begin{equation}
\left( \hat{H}, \hat{S}, \hat{V} \right) = \mathcal{H}\!\left( \hat{\mathbf{C}} \right)
\end{equation}
\begin{equation}
\hat{\theta} = \frac{\hat{H}}{179} \cdot 2\pi - \pi,\qquad\hat{m} = \frac{\hat{S}}{255} \cdot \sigma
\end{equation}
\begin{equation}
\hat{u} = \hat{m} \cos \hat{\theta},\qquad\hat{v} = \hat{m} \sin \hat{\theta}
\end{equation}
Flow quality is evaluated using the End-Point Error (EPE) over the valid pixel region $\Omega$:
\begin{equation}
\mathrm{EPE} = \frac{1}{|\Omega|} \sum_{p \in \Omega}\left\| \left( \hat{u}, \hat{v} \right)(p) - \left( u, v \right)(p) \right\|_2
\end{equation}

\section{Datasets and evaluation protocol}
\label{app:datasets}
\begin{figure*}[htbp]
    \centering
    \includegraphics[width=\linewidth]{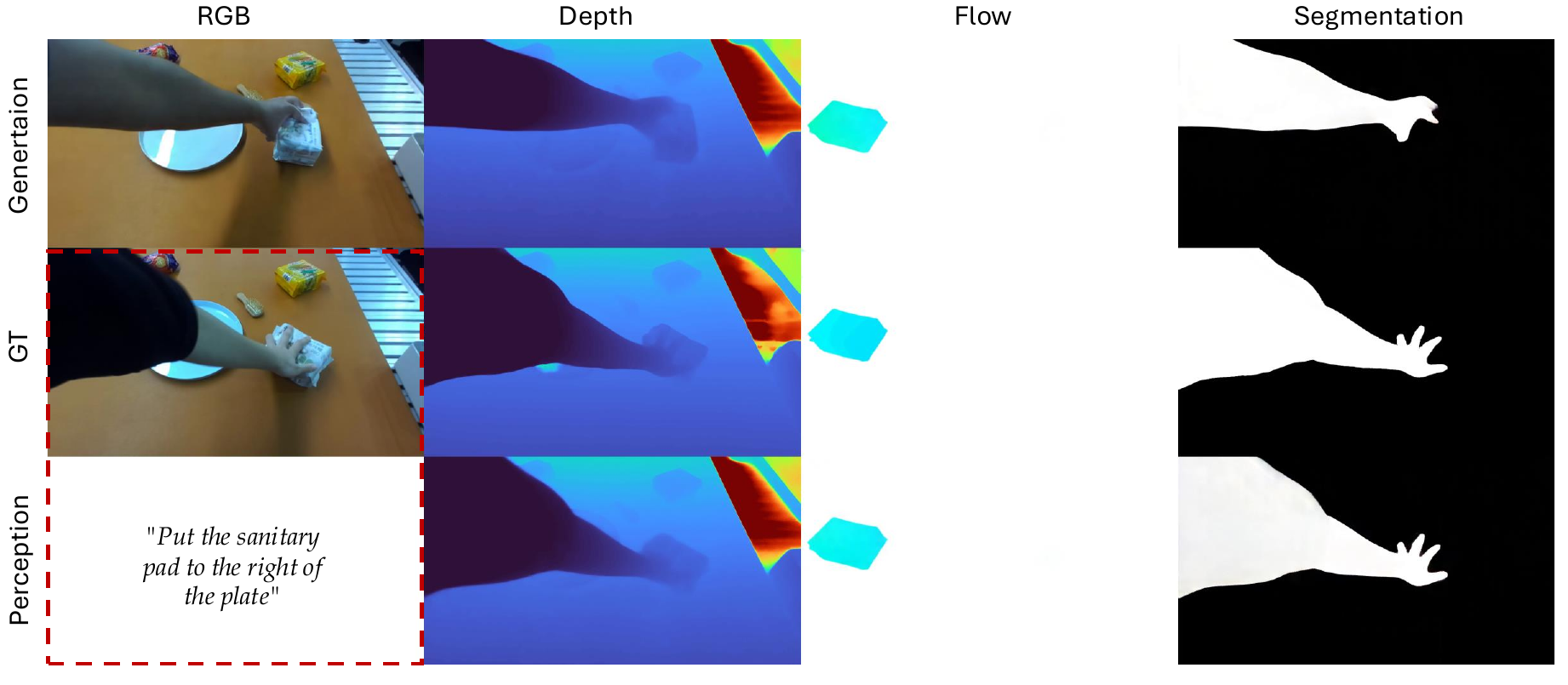} 
    \caption{\textbf{Qualitative Results on EgoTable (Human Hand)} Top: generative mode; middle: ground truth (GT); bottom: perception mode.}
    \label{fig:egotable}
\end{figure*}

\paragraph{Data sources.}
We combine RoboMIND~2.0~\citep{hou2025robomind2} episodes from UR5, Franka, Tianyi, and Tienkung with \emph{EgoTable}, an egocentric dataset collected for this study of human tabletop manipulation, as exemplified in Figure~\ref{fig:egotable}. It encompasses over 5,000 distinct tasks, featuring more than 200 manipulated objects, more than 100 reference objects, and 6 explicit spatial relations. For both RoboMIND 2.0 and EgoTable, each episode provides RGB video, synchronized sensor depth, and an action caption. We augment captions with domain tags specifying the data type, data source, and arm type, so generation can be conditioned on the input embodiment. Dataset statistics are shown in Table~\ref{tab:dataset_stats}.

\begin{table}[htbp]
\centering
\captionsetup{justification=centering,singlelinecheck=true}\caption{Dataset statistics across different sub-datasets.}
\label{tab:dataset_stats}
\small\begin{tabular}{llrr}
\toprule
\textbf{Category} & \textbf{Sub-dataset} & \textbf{Clips} & \textbf{Hours} \\ 
\midrule
\multirow{4}{*}{RoboMIND 2.0} 
 & UR5       & 48,459  &  359.42 \\
 & Tianyi    & 7,095   &  20.62 \\
 & Tienkung  & 6,831   &  42.42 \\
 & Franka    & 112,188 &  589.07  \\ 
\cmidrule(lr){2-4}
EgoTable & \textit{In-house Tasks} & 46,471 & 38.03   \\ 
\midrule
\multicolumn{2}{l}{\textbf{Total}} & \textbf{221,044} & \textbf{1049.58} \\ 
\bottomrule
\end{tabular}
\end{table}

\paragraph{Evaluation splits.}
For the training datasets, we reserve all episodes from $5\%$ of tasks for out-of-distribution evaluation and use the remaining $95\%$ of tasks for training. The RoboMIND 2.0 test set contains 600 episodes from 24 unseen real-world tasks. For real-to-simulation evaluation, we use RoboTwin~2.0~\citep{chen2025robotwin2}, comprising 300 episodes from 50 simulated tasks.

\section{Dense modality annotation}
\label{app:pseudolabels}
We annotate the dense modalities offline using sensor measurements and pretrained estimators. \emph{Depth:} Although the datasets include depth measurements from synchronized depth cameras, the raw maps contain invalid pixels. We apply Lingbot-Depth~\citep{tan2026lingbotdepth} to complete these regions and obtain dense depth maps.
\emph{Segmentation:} We use GroundingDINO~\citep{liu2024groundingdino} with embodiment-specific prompts (robot arm and gripper, or human hand) to localize the target. The detections initialize SAM2~\citep{ravi2025sam2} mask propagation across each clip, with periodic re-detection to mitigate drift.
\emph{Scene flow:} AllTracker~\citep{harley2025alltracker} estimates dense 2D displacement fields relative to the first frame. We filter the estimated fields according to the maximum displacement of each pixel over time to suppress outliers. These 2D fields can be combined with depth for 3D scene-flow evaluation.

\section{Flow evaluation metrics}
\label{app:flow_metrics}
We evaluate flow relative to the first frame. For 2D flow, endpoint error (EPE, pixels) is averaged over valid pixels. Let $e_i=\|\hat{\mathbf f}_i-\mathbf f_i\|_2$ and $g_i=\|\mathbf f_i\|_2$ be the error and ground-truth flow magnitude at valid pixel $i$. For generated flow, we also report normalized EPE, $\mathrm{nEPE}=\sum_i e_i/\sum_i g_i$, to normalize by the magnitude of ground-truth motion.
We additionally report $\mathrm{AAE}_{\mathrm{top}\,1\%}$, the mean angular error (in degrees) over the $1\%$ of valid pixels with the largest ground-truth flow magnitude. These metrics complement raw EPE, which can favor near-static predictions when reference motion is small.
For 3D scene flow, we report average and final trajectory errors (ADE and FDE, centimeters) in both absolute-position and first-frame-relative displacement forms. The displacement variants can partially cancel frame-to-frame depth fluctuations; we therefore emphasize absolute-position errors in the main comparison.

\section{Training details}
\label{app:details}

\paragraph{Backbone}
The backbone is a $36$-block diffusion transformer with hidden width $4096$, initialized from a Cosmos3-Nano \citep{nvidia2026cosmos3}; the visual tokenizer is a frozen Wan2.2 VAE ~\citep{wan2025wan22} with $(4, 16, 16)$ spatiotemporal compression, and text is tokenized by a Qwen3-VL-8B-Instruct tokenizer ~\citep{qwen3}.

\paragraph{Trainable and Frozen Components}
The backbone carries two parallel expert pathways per block, a reasoner and a generator, both initialized from Qwen3-VL-8B. We train the generator and freeze the reasoner throughout: within the transformer blocks, this splits the parameters into 396 trainable and 396 frozen tensors.

\paragraph{Resolution, Clip Length, and Token Budget}
Rather than fixing a batch size, samples are packed
sequentially into a fixed token budget of $12{,}288$ vision tokens per device per
step, drawing samples until the budget is filled. The token count of a clip is
\begin{equation}
N_{\mathrm{tok}} = \left\lfloor \frac{H}{32} \right\rfloor \cdot
\left\lfloor \frac{W}{32} \right\rfloor \cdot \left( 1 + \frac{T-1}{4} \right)
\end{equation}
where the spatial divisor is the product of the VAE spatial compression $16$ and the
patch size $2$. The budget therefore trades resolution against duration: it admits
$49$ frames at $704 \times 1280$, $149$ frames at $480 \times 640$, and $577$ frames
at $256 \times 320$. Because samples are packed until the token budget is reached, shorter clips allow more samples per step without reducing peak memory use. The token budget therefore controls memory consumption. Gradients are not accumulated.

\paragraph{Optimizer and Learning-Rate Schedule}
We use fused AdamW with $\beta = (0.9, 0.95)$, $\epsilon = 10^{-6}$, no weight decay,
and gradient-norm clipping at $1.0$. The peak learning rate is $8 \times 10^{-5}$.
The schedule is a linear warmup over the first $500$ steps from $0$ to the peak,
followed by a single cosine decay cycle scheduled over $8{,}000$ steps toward a
floor of $1 \times 10^{-5}$.

\paragraph{Dense-task Warmup Factor}
The loss weights of the three dense modalities are additionally 
ramped by a warmup factor $\lambda_m(n)$, which rises from $0$ to $1$ on a
cosine over the first $10\%$ of the scheduled run, \textit{i.e.},\ $n_{\lambda} = 800$
steps:
\begin{equation}
\lambda_m(n) = \tfrac{1}{2}
\left( 1 - \cos \pi \min\!\left( \frac{n}{n_{\lambda}}, 1 \right) \right)
\end{equation}
The video branch is never scaled by $\lambda$. This schedule is intended to preserve the pretrained image-to-video capability during early optimization. 

\paragraph{Compute Budget}
Training ran on $64$ NVIDIA A800 GPUs under hybrid
sharded data parallelism. Each optimizer step takes $37.5$\,s wall clock, with most of the time spent evaluating the four task-specific pathways in the later transformer blocks. The run consumed $85$\,hours for
$8{,}000$ steps, approximately $5{,}400$ GPU-hours.

\section{Ablation settings and detailed results}
\label{app:ablation_details}
\paragraph{Protocol.}
All ablation models are trained exclusively on the $95\%$ training-task split of RoboMIND~2.0 UR5 and evaluated on the remaining $5\%$: eight unseen tasks with 25 episodes each, for 200 test episodes. We evaluate checkpoints after 3,000 training steps. Unless varied, the task-agnostic prefix contains 16 transformer blocks and the prediction noise level is $\sigma=0.6$. The sharing and noise sweeps use equal modality weights. The modality-composition study assigns a weight of 0.25 to each active dense objective and 1 to TI2V.

\paragraph{Number of task-agnostic blocks.}
Figure~\ref{fig:ablation_studies} shows that the optimal degree of task-agnostic sharing varies across metrics, with no single prefix length dominating all tasks. As the task-agnostic prefix becomes too long, both video generation and dense perception exhibit a broad performance degradation, suggesting that excessive sharing limits modality-specific adaptation. An intermediate prefix provides the best overall trade-off across video quality, depth, segmentation, and flow. We therefore adopt a balanced intermediate configuration, supporting the design choice of preserving task-adaptive capacity in the later blocks.

\paragraph{Dense-readout noise.}
Figure~\ref{fig:ablation_studies} evaluates a fixed checkpoint across different prediction-noise levels. The video metrics remain unchanged because the sweep only alters the dense readout state. In contrast, the dense modalities exhibit distinct trends: depth improves toward relatively high noise levels before degrading at the extreme end of the sweep; segmentation performs best at lower noise levels and gradually deteriorates as the noise increases; and flow improves over most of the range before saturating and slightly reversing at the highest noise level. These modality-dependent trends motivate using separate readout noise levels for depth, segmentation, and flow.

\section{Additional results}
\label{app:qualitative}

\begin{figure*}[htbp]
    \centering
    \includegraphics[width=\linewidth]{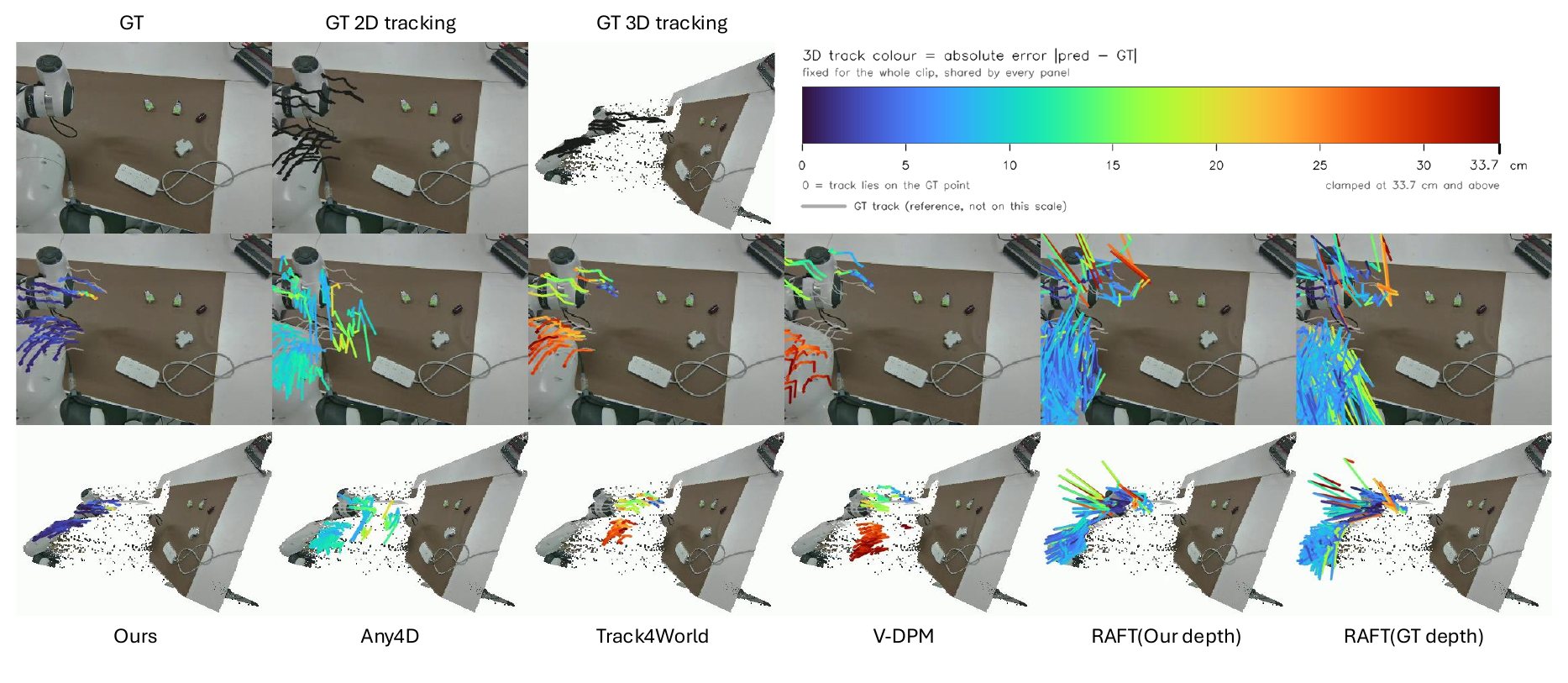} 
    \caption{\textbf{Qualitative 3D tracking results.} We obtain 3D trajectories by combining estimated depth and 2D optical flow. The visualizations compare trajectory alignment in image and 3D views. Tracking4World~\cite{lu2026track4world} aligns with the ground truth in the image plane but exhibits a visible depth offset in the side view. Temporal fluctuations in estimated depth can also affect displacement-based trajectory metrics. Trajectory color indicates the magnitude of the absolute error.} 
    \label{fig:tracking_3d}
\end{figure*}

\begin{figure*}[htbp]
    \centering
    \includegraphics[width=0.9\linewidth]{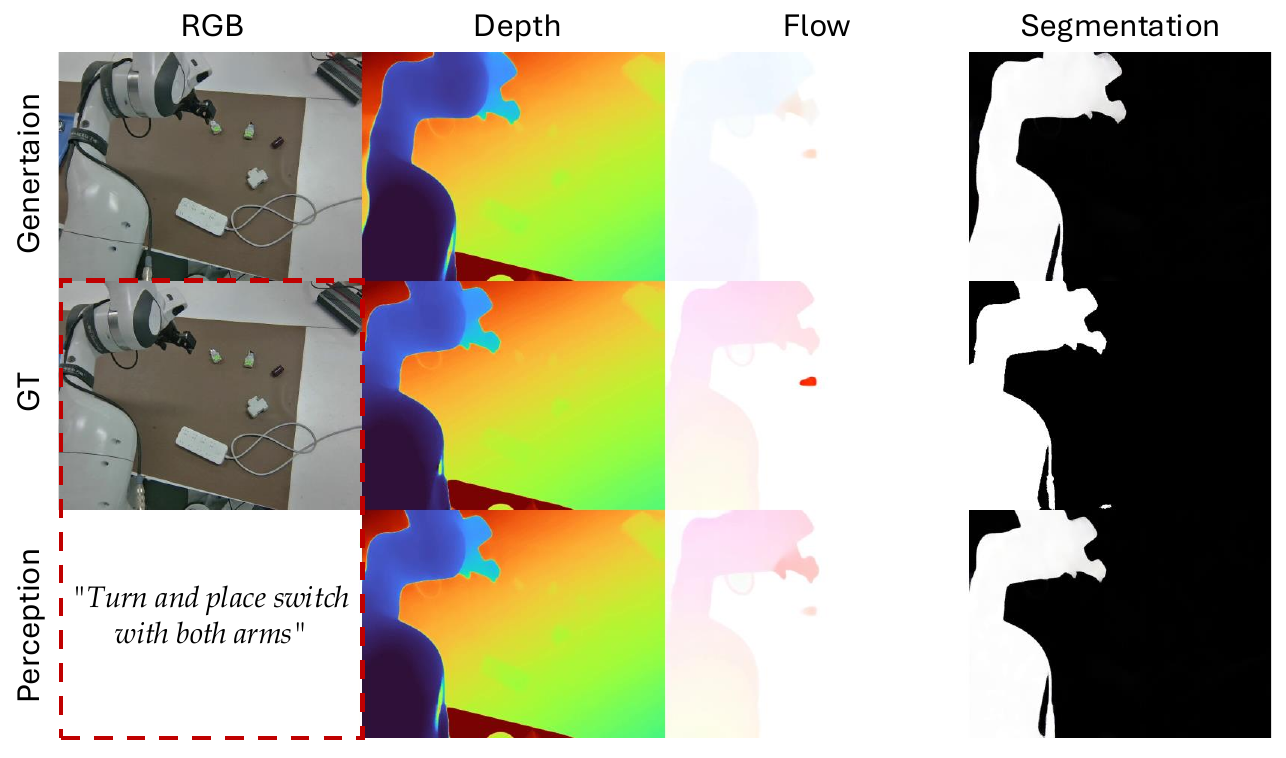} 
    \caption{\textbf{Qualitative Results on RoboMIND 2.0 (Franka Embodiment)} Top: generative mode; middle: ground truth (GT); bottom: perception mode.} 
    \label{fig:franka}
\end{figure*}

\begin{figure*}[htbp]
    \centering
    \includegraphics[width=\linewidth]{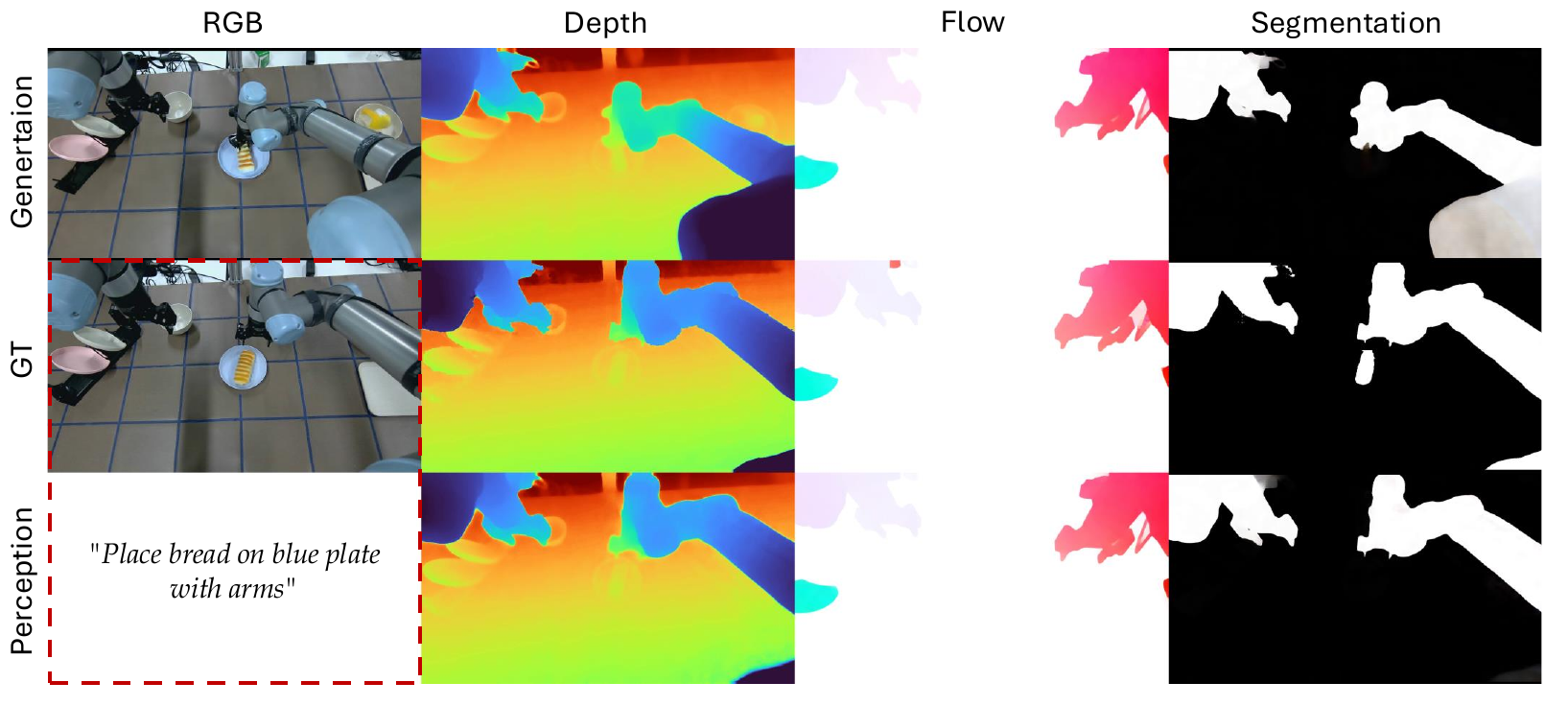} 
    \caption{\textbf{Qualitative Results on RoboMIND 2.0 (UR5 Embodiment)} Top: generative mode; middle: ground truth (GT); bottom: perception mode.} 
    \label{fig:ur5}
\end{figure*}

\begin{figure*}[htbp]
    \centering
    \includegraphics[width=\linewidth]{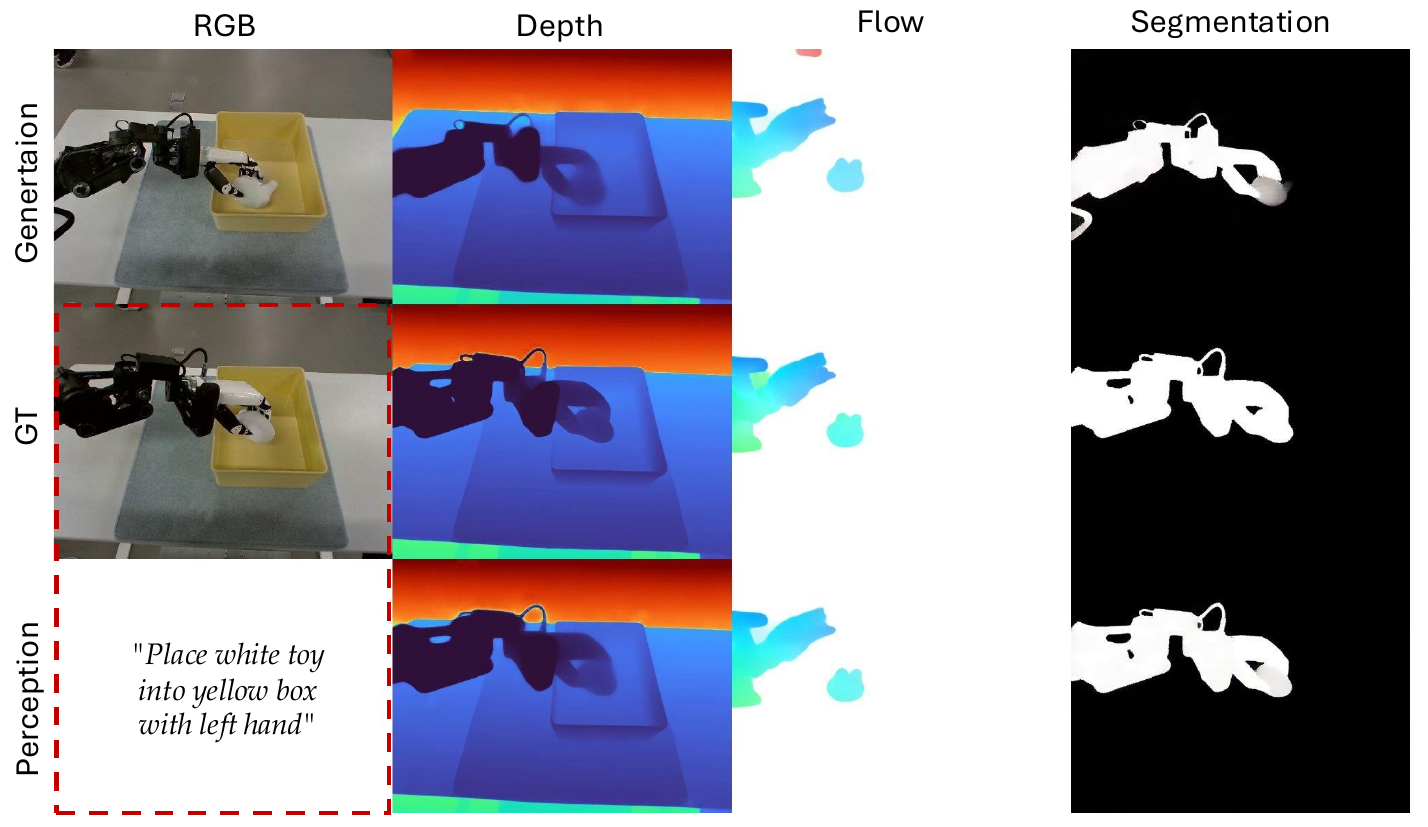} 
    \caption{\textbf{Qualitative Results on RoboMIND 2.0 (Tianyi Embodiment)} Top: generative mode; middle: ground truth (GT); bottom: perception mode.} 
    \label{fig:tianyi}
\end{figure*}

\begin{figure*}[htbp]
    \centering
    \includegraphics[width=\linewidth]{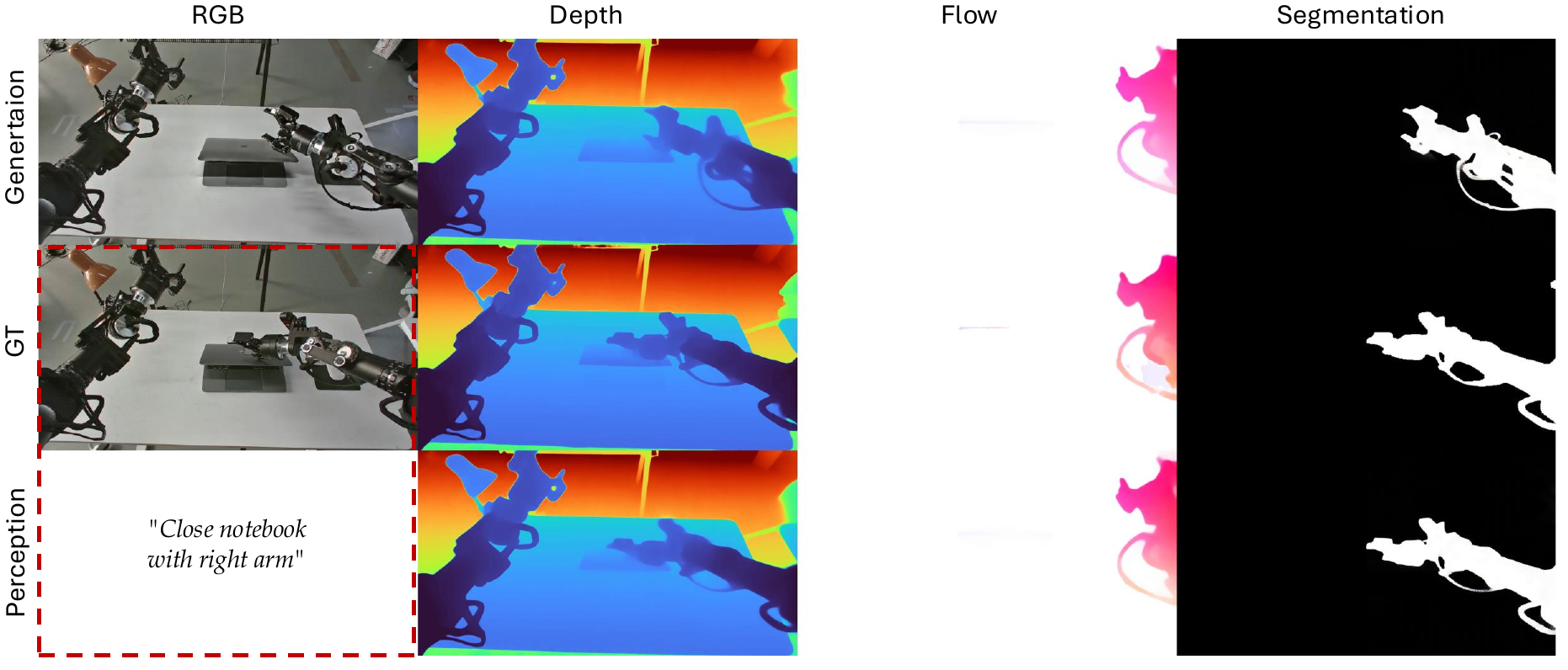} 
    \caption{\textbf{Qualitative Results on RoboMIND 2.0 (Tienkung Embodiment)} Top: generative mode; middle: ground truth (GT); bottom: perception mode.} 
    \label{fig:tienkung}
\end{figure*}

\begin{figure*}[htbp]
    \centering
    \includegraphics[width=0.9\linewidth]{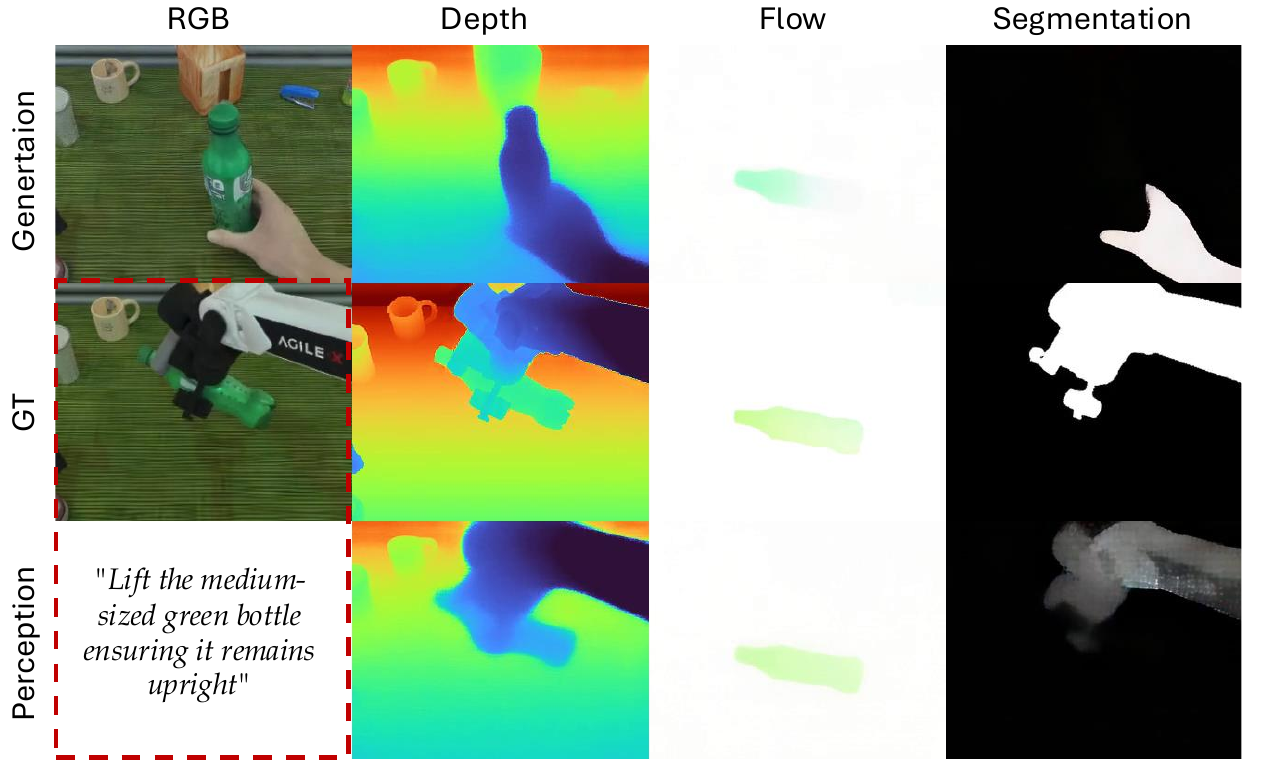} 
    \caption{\textbf{Qualitative Results on RoboTwin 2.0 (Simulated AgileX Embodiment)} Top: generative mode; middle: ground truth (GT); bottom: perception mode.}
    \label{fig:robotwin_2.0}
\end{figure*}

\end{document}